%% file: root.tex
\documentclass[letterpaper, 10 pt, journal, twoside]{IEEEtran}
\usepackage{cite}
\usepackage{amsmath,amssymb,amsfonts}
\usepackage{algorithmic}
\usepackage{graphicx}
\usepackage{textcomp}
\usepackage{xcolor}

\usepackage{bm}
\usepackage[binary-units = true]{siunitx}
\usepackage{multirow}
\usepackage{tikz}
\usetikzlibrary{arrows}
\usepackage[colorlinks=true,allcolors=blue,urlcolor=blue]{hyperref} %needs to be the last import

\DeclareSIUnit\px{px}

\newcommand{\secref}[1]{Section~\ref{#1}}
\newcommand{\figref}[1]{Fig.~\ref{#1}}
\newcommand{\tabref}[1]{Table~\ref{#1}}

\newcommand{\etal}{\emph{et al.}}
\newcommand{\ie}{\emph{i.e.}}
\newcommand{\eg}{\emph{e.g.}}

\newcommand{\RR}{{\rm I\!R}}
\newcommand{\Lagr}{\mathcal{L}}

\newcommand{\norm}[1]{\left\lVert#1\right\rVert}

\DeclareMathOperator*{\argmin}{arg\,min}
\DeclareMathOperator*{\sign}{sign}
\DeclareMathOperator*{\trans}{trans}

\include{tables}

\begin{document}

\thispagestyle{empty}
\onecolumn

\hfill \vspace{-0.5em}
\begin{center}
	This paper has been accepted for publication in \emph{IEEE Robotics and Automation Letters}.
	\vspace{2em}
	
	DOI: \href{https://doi.org/10.1109/LRA.2019.2922618}{10.1109/LRA.2019.2922618} \\
	%IEEE Xplore:
	\vspace{3em}
\end{center}

\textcopyright\ 2019 IEEE. Personal use of this material is permitted. Permission from IEEE must be obtained for all other uses, in any current or future media, including reprinting/republishing this material for advertising or promotional purposes, creating new collective works, for resale or redistribution to servers or lists, or reuse of any copyrighted component of this work in other works.
\clearpage

\twocolumn
\setcounter{page}{1}

\title{Automatic Multi-Sensor Extrinsic Calibration for Mobile Robots}

%\author{\IEEEauthorblockN{David Zu\~niga-No\"el, Jose-Raul Ruiz-Sarmiento, Ruben Gomez-Ojeda, and Javier Gonzalez-Jimenez\textsuperscript{*}}}
\author{David Zu\~niga-No\"el, Jose-Raul Ruiz-Sarmiento, Ruben Gomez-Ojeda, and Javier Gonzalez-Jimenez$^{*}$
\thanks{Manuscript received: February 24, 2019; Revised: April 10, 2019; Accepted: May 29, 2019.} %Use only for final RAL version
\thanks{This paper was recommended for publication by Editor Dezhen Song upon evaluation of the Associate Editor and Reviewers' comments. This work was supported by the research projects \emph{WISER} (DPI2017-84827-R), funded by the Spanish Government and financed by the European Regional Development's funds (FEDER), \emph{MoveCare} (ICT-26-2016b-GA-732158), funded by the European H2020 program, the European Social Found through the Youth Employment Initiative for the promotion of young researchers, and by a contract from the {I-PPIT} program of the University of Malaga.} % Use only for final RAL version
\thanks{$^{*}$The authors are with the Machine Perception and Intelligent Robotics group (MAPIR), System Engineering and Automation Department, and the Biomedical Research Institute of Malaga (IBIMA), University of Malaga, Spain. Corresponding author: {\tt\footnotesize dzuniga@uma.es}}
\thanks{Digital Object Identifier (DOI): see top of this page.} % Use only for final RAL version
}

\maketitle

\begin{abstract}
In order to fuse measurements from multiple sensors mounted on a mobile robot, it is needed to express them in a common reference system through their relative spatial transformations. In this paper, we present a method to estimate the full 6DoF extrinsic calibration parameters of multiple heterogeneous sensors (Lidars, Depth and RGB cameras) suitable for automatic execution on a mobile robot. Our method computes the 2D calibration parameters (x, y, yaw) through a motion-based approach, while for the remaining 3 parameters (z, pitch, roll) it requires the observation of the ground plane for a short period of time. What set this proposal apart from others is that: i) all calibration parameters are initialized in closed form, and ii) the scale ambiguity inherent to motion estimation from a monocular camera is explicitly handled, enabling the combination of these sensors and metric ones (Lidars, stereo rigs, etc.) within the same optimization framework.
%Additionally, outlier observations arising from local sensor drift are automatically detected and removed from the calibration process.
We provide a formal definition of the problem, as well as of the contributed method, for which a C++ implementation has been made publicly available. The suitability of the method has been assessed in simulation an with real data from indoor and outdoor scenarios. Finally, improvements over state-of-the-art motion-based calibration proposals are shown through experimental evaluation.
\end{abstract}

% Keywords appear just beneath the abstract. Use only for final RAL version. 

%--------------------------------------------------------
%                    1. Introduction
%--------------------------------------------------------

\section{Introduction} 
% Background
% In this part you have to make clear what the context is. Ideally, you should give an idea of the state-of-the art of the field the report is about. But keep it short.

% Drop letter for first word of the Introduction. Use only for final RAL version.
\IEEEPARstart{A}{utonomous} mobile robots require sensors to perceive the environment and estimate their state. Combining measurements from different sensors is a need to improve robustness and to compensate for individual sensor limitations. In order to express measurements into a common reference system, accurate relative transformations between the sensors are required (\ie\ extrinsic calibration). For example, when doing Visual Odometry~\cite{visual_odometry} it is a common practice to consider also information from an odometer~\cite{analytical_odometer-camera_calibration, planar-odometry_calibration, simultaneous_odometry-camera_calibration}, which requires knowing the spatial transformation between the camera and the wheel odometry system.

Two main approaches can be found in the literature for extrinsic calibration: i) exploiting \emph{a priori} knowledge about the environment or ii) relying on per-sensor motion estimates. In the first case, \emph{a priori} information is typically in the form of scene patterns~\cite{laser-camera_corner_calibration, rgbd_sphere_calibration, depth-sensor_planar_calibration} or known landmarks~\cite{simultaneous_odometry-camera_calibration}. 
Pattern-based calibration methods are usually developed for specific pairs of sensors. For example: a camera and a 2D laser scanner~\cite{laser-camera_corner_calibration}, or monocular and depth cameras~\cite{rgbd_sphere_calibration}. Using specific calibration techniques to cover all sensors pairs in a multi-sensor configuration makes the calibration process complex and almost impractical. Additionally, relying on known landmarks for calibration prevents the automatic execution of the calibration process since they have to be placed in the environment beforehand. However, automatic calibration methods are interesting for mobile robots: they demand re-calibration capabilities for long-term operation since the relative transformation between their sensing devices can change due to crashes, vibrations, human intervention, etc~\cite{motion-based_calibration_w_time, ruiz2017building}.

In contrast to pattern-based calibration methods, motion-based ones do not need to modify the environment and they can be used with any sensing devices, provided that ego-motion can be inferred from them. Monocular cameras are the exception, being excluded due to the scale ambiguity problem, which has to be considered explicitly. Another practical limitation arises from the planar movement of mobile robots, which prevents the estimation of the full 6DoF extrinsic calibration parameters solely form incremental motions, as shown in~\cite{analytical_odometer-camera_calibration}. 

%\red{it is typically assumed some kind of \emph{a priori} knowledge about the environment in the form of, for example, known landmark poses~\cite{simultaneous_odometry-camera_calibration} or scene patterns~\cite{laser-camera_corner_calibration, rgbd_sphere_calibration, depth-sensor_planar_calibration}. However, these requirements can prevent the automatic execution of the calibration, which is required by mobile robots working in human environments where the relative transformations between sensing devices can change due to crashes, vibrations, human intervention, etc. To address this, calibration methods based on sensor egomotion are a good option since any \emph{a priori} knowledge is required. Moreover, they can work with any sensor from which egomotion can be inferred. Monocular cameras are usually the exception, being excluded due to the scale ambiguity appearing in their reconstructed motions, which must be considered explicitly.}

\begin{figure}[t]
    \centering
\includegraphics[width=0.47\textwidth]{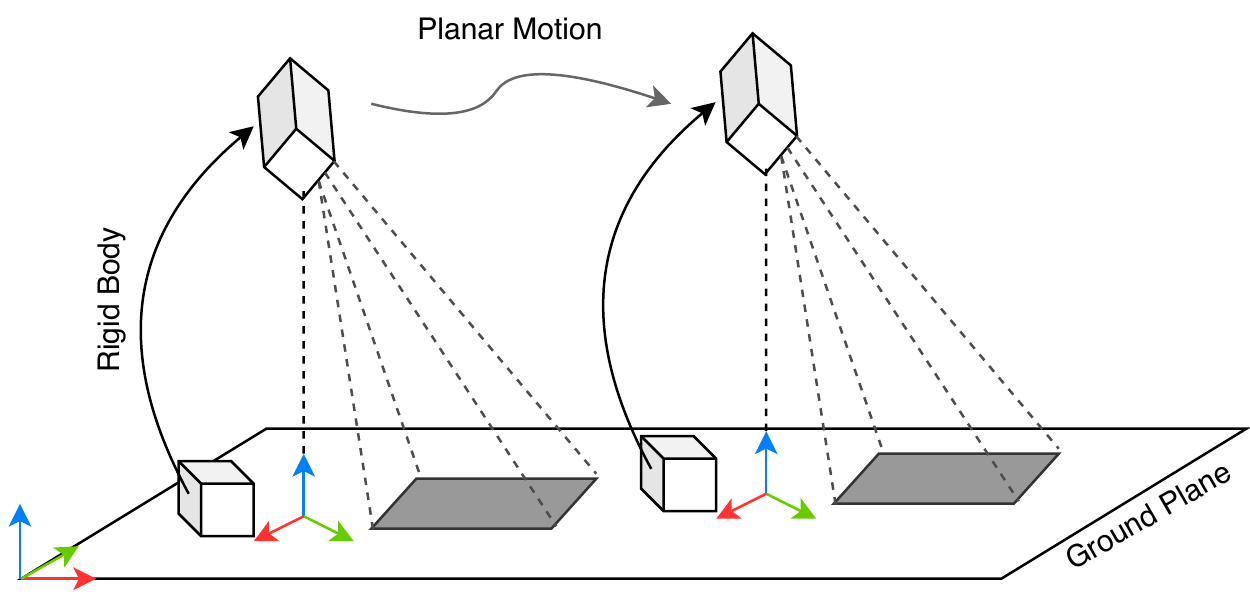}
    \caption{A generic two-sensor setting rigidly attached to a robot (omitted for clarity) moving on a planar surface, with a main sensor for planar motion estimation (\eg~an odometer) drawn on the ground plane, and an additional sensor observing the plane (\eg~a camera). The estimation of the parameters that defines the constant rigid body motion connecting both sensors is the goal of our calibration method. In case of using a monocular camera, the scale of the reconstructed motion is also estimated.}
    \label{fig:two-sensor_calibration}
    %\vspace{-0.2cm}
\end{figure}

In this paper, we contribute a method to extrinsically calibrate multiple heterogeneous sensors such as odometers, visual and depth cameras, 2D and 3D laser scanners, etc., mounted on a robot moving on a planar ground (see \figref{fig:two-sensor_calibration}). Our approach computes the 6 extrinsic parameters in two main steps: the 2D calibration parameters (\emph{x}, \emph{y}, \emph{yaw}) are obtained from per-sensor incremental motions, while the remaining 3 parameters (\emph{z}, \emph{pitch}, \emph{roll}) are computed from the observation of the ground plane. The ground plane has to be observed for a short period of time, and this requisite is fulfilled by most robotic sensor configurations. By decoupling the extrinsic calibration problem into two simpler sub-problems we are able to solve all parameters in closed form, thus removing the necessity to provide initial estimates. Another important aspect of our method is that it formulates the calibration problem taking into account the scale ambiguity in monocular cameras, which is estimated in the motion-based step. Thus, our approach allows to combine these commonly-used devices with other types of sensors in the same calibration framework. A C++ open source implementation of the contributed method is publicly available for download at~\url{https://github.com/dzunigan/robot_autocalibration}.

Our proposal has been validated in a simulation experiment and assessed with real data from two different datasets: one in outdoor scenarios containing information from a monocular camera and a GPS~\cite{dataset_cm_accuracy_groundtruth}, and another one gathered indoors using one of our mobile robotic platforms, equipped with an odometer, a monocular camera and a 2D laser scanner~\cite{giraff}. We also report improvements in terms of accuracy and consistency of the estimates over the state-of-the-art motion-based approach from Della Corte \etal~\cite{motion-based_calibration_w_time}.

\section{Related Work}

Most extrinsic calibration methods found in the literature consider specific sensors and exploit some \emph{a priori} information about the observed scene. For example, the work in~\cite{laser-camera_corner_calibration} uses scene corners (orthogonal trihedrons) to calibrate a 2D laser scanner and a camera. The information provided by a spherical object is used in~\cite{rgbd_sphere_calibration} to calibrate a camera and a depth sensor. The overlap requirement between the sensors' field of view is relaxed in~\cite{depth-sensor_planar_calibration} to calibrate multiple depth cameras from the common observation of planar structures. In contrast to these methods, \emph{a priori} knowledge is not required by motion-based calibration approaches, and hence can be used automatically in unstructured environments.

% From egomotion
The extrinsic calibration from individual sensor egomotion was first considered, to the best of our knowledge, by Brookshire and Teller in~\cite{coplanar_calibration}. They present an observability analysis of the 2D extrinsic calibration problem and a solution based on the iterative minimization of a least-squares cost function. Similarly, in~\cite{odometry-sensor_calib}, Censi~\etal\ consider the calibration of both, a generic sensor (for which egomotion can be estimated) with respect to the robot reference system, and the intrinsic parameters of a differential drive configuration. They also analyze the observability of the calibration problem and propose a closed form solution to a least-squares formulation. In~\cite{per-sensor_egomotion_calibration}, Brookshire and Teller extend their previous formulation by considering the calibration of the 3D transformation parameters. Schneider~\etal\ present in~\cite{odometry-based_calibration} an online extrinsic calibration approach based on the Unscented Kalman Filter.% The  is not considered in these works.

Unlike these previous works, the authors in~\cite{analytical_odometer-camera_calibration} address the scale ambiguity of monocular cameras proposing a closed form least-squares solution to the odometer-camera calibration problem based on incremental motions. Their approach does not require any \emph{a priori} information about the environment, and thus allows for an automatic execution. Zienkiewicz and Davison consider in~\cite{planar-odometry_calibration} a similar calibration problem. They propose a method based on the parameterization of the homography. The solution is found by minimizing the photometric errors induced by the homography arising from planar motions observing the ground plane.

%Multisensor calibration
All aforementioned works are limited to the calibration of pairs of sensors. In contrast, the work in~\cite{camodocal} considers the calibration of multiple cameras with an odometer. Their pipeline estimates the intrinsic camera parameters as well as the extrinsic parameters with respect to an odometer (initialized from~\cite{analytical_odometer-camera_calibration}), but it is designed to work only with imaging sensors. The work in~\cite{gauss-helmert_calibration} tackles the calibration of multiple sensors explicitly. Their formulation is based on the Gauss-Helmert model, where all motions have to be expressed in the same scale, an thus it is not suitable for cameras. Taylor and Nieto propose in~\cite{multiple_calibration} an algorithm to extrinsically calibrate multiple sensors from 3D motions, including cameras, by solving multiple hand-eye calibration problems. Recently, Della Corte~\etal\ presented in~\cite{motion-based_calibration_w_time} an extrinsic calibration framework from 3D motions, where each sensor's time delay are also estimated. However, as shown in~\cite{analytical_odometer-camera_calibration}, the full 6DoF calibration parameters cannot be estimated solely from planar motions performed by mobile robots.
% Maybe: with a downward-looking camera, (Extrinsics Autocalibration for Dense Planar Visual Odometry)

In contrast to previous approaches, the method presented in this paper estimates the 6DoF calibration parameters of multiple, heterogeneous sensors (including monocular cameras) mounted on a mobile robot performing planar motions. For that, it relies on incremental sensor motions as well as the observation of the ground plane. The \emph{a priori} information requirement introduced by the calibration step based on the ground plane detection is soft enough to allow automatic execution with different sensors (\eg\ monocular cameras or 3D laser scanners) without requiring overlapping field of view configurations. Additionally, all extrinsic calibration parameters are estimated in closed form, and hence initial guess values are not required. As a consequence, our method is suitable for automatic execution by mobile robots.

\section{Problem Formulation} \label{sec:Formulation}

%--------------------------------------------------------
%              3.A Motion-based Calibration
%--------------------------------------------------------
\subsection{Motion-based Calibration Coplanar Sensors} \label{sec:two_coplanar_formulation}

First, we consider the motion-based extrinsic calibration of two coplanar sensors under the assumptions that: i) 2D synchronized incremental poses are available for each of them, and ii) the translation component may be expressed in different scales. Our objective is to estimate the fixed 2D transformation between the two sensors that better explains the difference in the observed incremental motions.

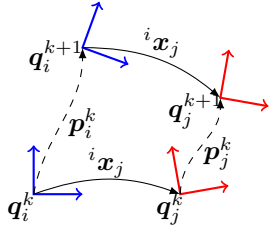
\begin{figure}[t]
    \centering
    \begin{tikzpicture}[scale=0.65]
    %Reference systems
    \draw [<->, thick, blue] (0,1) -- (0,0) -- (1,0);
    \draw [<->, thick, red] (2.8264,0.9848) -- (3,0) -- (3.9848,0.1736);
    \draw [<->, thick, blue] (1.3420,3.9397) -- (1,3) -- (1.9397,2.6580);
    \draw [<->, thick, red] (3.9927,2.9587) -- (3.8191,1.9739) -- (4.8039,1.8003);
    %Labels
    \node at (-0.25,-0.25) {$\bm{q}^k_i$};
    \node at (2.7972,-0.2896) {$\bm{q}^k_j$};
    \node at (0.4574,2.8250) {$\bm{q}^{k+1}_i$};
    \node at (3.3152,1.7074) {$\bm{q}^{k+1}_j$};
    %Curves
    \draw [-latex, dashed] (0,0) to [out=80,in=270] (1,3);
    \draw [-latex, dashed] (3,0) to [out=90,in=270] (3.8191,1.9739);
    \draw [-latex] (0,0) to [out=20,in=160] (3,0);
    \draw [-latex] (1,3) to [out=0,in=140] (3.8191,1.9739);
    %Labels
    \node at (1.0119,1.4547) {$\bm{p}^k_i$};
    \node at (3.7790,0.8336) {$\bm{p}^k_j$};
    \node at (1.5,0.65) {$^i\bm{x}_j$};
    \node at (2.6319,3.0978) {$^i\bm{x}_j$};
    \end{tikzpicture}
    \caption{Extrinsic calibration of the $i$-th and $j$-th sensors from incremental motions. The observed incremental motions $\bm{p}^k_i, \bm{p}^k_j \in \text{SE(2)}$ at time step $k$ are related by the fixed relative transformation between the two sensors. We consider the calibration parameters between sensors $^i\bm{x}_j \in \text{Sim(2)}$ as a rigid body motion followed by a scaling.}

    \label{fig:coplanar_sensor_calibration}
    %\vspace{-0.2cm}
\end{figure}

More formally, our aim is to estimate the parameters of the 2D similarity transformation between the $i$-th and the $j$-th sensors $^i\bm{x}_j = \left( x_x, x_y, x_\theta, x_s \right) \in \text{Sim(2)}$, where $x_x, x_y \in \RR$ are the translation components, $x_\theta \in \RR$ the rotation angle, $x_s \in \RR^+$ the scaling factor (see \figref{fig:coplanar_sensor_calibration}), and $\text{Sim(2)}$ the group of orientation-preserving similarity transformations\footnote{Similarity transformations are rigid transformations followed by a scaling~\cite{sim(n)}.}. The calibration transformation expresses the measurements taken from the \mbox{$j$-th} sensor $^j\bm{m} \in \RR^2$ into the $i$-th reference frame as~\cite{sim(n)}:
\begin{equation} \label{eq:sim2_transformation}
    ^i\bm{m} = x_s \left(R(x_\theta)\,{^j\bm{m}} + \begin{bmatrix}
    x_x\\
    x_y
    \end{bmatrix}\right),
\end{equation}
where $R(\theta)$ is the 2D rotation matrix defined by the angle $\theta$:
\begin{equation}
    R(\theta) \triangleq \begin{bmatrix} \cos\theta & -\sin\theta \\
    \sin\theta & \cos\theta
    \end{bmatrix}.
\end{equation}

% and similar to~\cite{odometry-sensor_calib}
From \eqref{eq:sim2_transformation}, we can define the group operator of Sim(2) as:
\begin{equation}
    \begin{split}
    \oplus\colon & \text{Sim(2)} \times \text{Sim(2)} \to \text{Sim(2)}\\
                 & \bm{a} \oplus \bm{b} = \begin{bmatrix} {b_s}^{-1}a_x + b_x\cos(a_\theta) - b_y\sin(a_\theta)\\
                 {b_s}^{-1}a_y + b_x\sin(a_\theta) + b_y\cos(a_\theta)\\
                 a_\theta + b_\theta\\
                 a_s b_s
                 \end{bmatrix}
\end{split},
\end{equation}

and its inverse as:
\begin{equation}
    \begin{split}
    \ominus\colon & \text{Sim(2)} \to \text{Sim(2)}\\
                  & \ominus\,\bm{a} = \begin{bmatrix}- a_s a_x\cos(a_\theta) - a_s a_y\sin(a_\theta)\\
                  a_s a_x\sin(a_\theta) - a_s a_y\cos(a_\theta)\\
                  -a_\theta\\
                  {a_s}^{-1}
    \end{bmatrix}.
    \end{split}
\end{equation}

We consider incremental rigid body motions as input for the calibration process, expressed as $\bm{p}^k_i \in \text{SE(2)}$ between sample times $k$ and $k+1$. Notice that a SE(2) transformation is a particular case of Sim(2) with a scaling factor set to the identity, thus the previously defined operators for Sim(2) also hold for SE(2) transformations.
%The Sim(2) transformations can be seen as a generalization of the SE(2) rigid body motions. In such case, a SE(2) transformation can be expressed in Sim(2) with scaling factor set to the identity, where the previously defined operators for Sim(2) also hold for SE(2) transformations.

The incremental motions can be derived from the sensor poses $\bm{q}^k_i, \bm{q}^{k+1}_i \in \text{SE(2)}$ as\footnote{Throughout this paper, we set to $\ominus$ a higher precedence than $\oplus$ to improve the readability of expressions involving $\text{Sim(2)}$ operations.}:
\begin{equation} \label{eq:incremental_motion}
    \bm{p}^k_i \triangleq \ominus\, \bm{q}^k_i \oplus \bm{q}^{k+1}_i.
\end{equation}

The poses corresponding to the $j$-th sensor can be expressed in terms of the $i$-th sensor poses and the extrinsic calibration between them:
\begin{equation}
    \bm{q}^k_j = \bm{q}^k_i \oplus\,\!^i\bm{x}_j
\end{equation}
and, therefore, its incremental motion is given by:
\begin{equation} \label{eq:relation}
    \bm{p}^k_j = \ominus\, \left(\bm{q}^k_i \oplus\,\!^i\bm{x}_j\right) \oplus \left(\bm{q}^{k+1}_i \oplus\,\!^i\bm{x}_j\right).
\end{equation}

Rearranging terms in \eqref{eq:relation} and applying \eqref{eq:incremental_motion}, the relative motions $\bm{p}^k_i, \bm{p}^k_j \in \text{SE(2)}$ for the $i$-th and $j$-th sensors at a given time step $k$ and the extrinsic calibration parameters $^i\bm{x}_j \in \text{Sim(2)}$ are related by:
\begin{equation} \label{eq:relation_simplified}
    \bm{p}^k_j = \ominus\, ^i\bm{x}_j \oplus \bm{p}^k_i \oplus\,\!^i\bm{x}_j.
\end{equation}

This relationship allows us to establish a cost function for a least-squares formulation of the two-sensor extrinsic calibration problem:
\begin{gather}
    \label{eq:pairwise_ls}
    ^i\bm{x}^*_j = \argmin_{\bm{x}}\, \mathcal{C}_{ij}(\bm{x})\\
    \label{eq:cost_definition}
    \mathcal{C}_{ij}(\bm{x}) \triangleq \frac{1}{2}\sum_k \norm{\bm{\epsilon}^k_{ij}(\bm{x})}^2_2\\
    \label{eq:error_terms_definition}
    \bm{\epsilon}^k_{ij}(\bm{x}) \triangleq \bm{p}^k_j - \ominus\, \bm{x}\oplus\bm{p}^k_i\oplus \bm{x},
\end{gather}
where the error function in~\eqref{eq:error_terms_definition} refers to the relationship between the observed incremental motions and the calibration parameters in~\eqref{eq:relation_simplified}.

%--------------------------------------------------------
%           3.B Sensor Coplanarity Relaxation
%--------------------------------------------------------

\subsection{Sensor Coplanarity Relaxation} \label{sec:coplanarity_relaxation_formulation}
Now, consider the case of a sensor in a generic configuration, under the assumption that the plane in which motions are performed can be observed. Our objective is to set a common reference system to estimate the relative pose parameters in order to project estimated motions into this plane, and hence enforcing the coplanarity constraint.

Formally, we want to estimate the relative rigid body motion $(\bm{R}_i, \bm{t}_i) \in \text{SE(3)}$ between the $i$-th sensor and the ground plane. For convenience, we set the plane's local reference system to be at the projection of the sensor's origin into the plane and with its $z$-axis parallel to the plane's normal pointing upwards (see \figref{fig:two-sensor_calibration}). Note that the in-plane rotation (around the $z$-axis) can be arbitrarily set as it will be calibrated latter on (as the $x_\theta$ parameter).
%\blue{For simplicity, we set it to be aligned with the reference sensor's orientation.}

Thus, ignoring the relative in-plane position and orientation, the remaining 3DoF parameters $^i\bm{x} = \left( x_z, x_\alpha, x_\beta \right) \in \RR^3$ for each sensor are the target of our estimation at this step. The parameter $x_z \in \RR^+$ represents the perpendicular distance and $x_\alpha, x_\beta \in \RR$ two rotation angles. More specifically, we define the relative SE(3) transformation to be:
\begin{equation} \label{eq:se3_parameters}
    \bm{R}_i \triangleq R_y(x_\beta)R_x(x_\alpha),\quad \bm{t}_i \triangleq \begin{bmatrix} 0 & 0 & x_z \end{bmatrix}^\top
\end{equation}
where $R_x(\cdot)$ and $R_y(\cdot)$ represent parameterized 3D rotation matrices along the $x$ and $y$ axes, respectively.

%\begin{figure}[t]
%    \centering
%    \includegraphics[width=0.9\columnwidth]{figures/GroundCalibrationDiagram.png}
%    \caption{A generic sensor observing the ground plane and its projected local reference coordinates. The relative rigid body transformation between the sensor and the local ground reference system remains constant with inplane motions. The estimation of the parameters defining this constant rigid body motion is the goal of the coplanarity relaxation.}
%    \label{fig:ground-plane_calibration}
    
%    \vspace{-0.3cm}
    
%\end{figure}

% The observed points and plan are related [...] and this relationship allows us to define ls problem
Hence, the $j$-th 3D point $^i\bm{m}_j \in \RR^3$ observed by the $i$-th sensor can be expressed in the local ground coordinates by applying the relative rigid body transformation as:
\begin{equation} \label{eq:planar_relation}
    ^p\bm{m}_j = \bm{R}_i\,^i\bm{m}_j + \bm{t}_i
\end{equation}

Based on this relationship, we can define a cost function for a weighted least-squares formulation of the coplanarity relaxation problem:
\begin{gather}
    \label{eq:coplanarity_ls}
    ^i\bm{x}^* = \argmin_{\bm{x}}\, \mathcal{E}_i(\bm{x})\\
    \label{eq:coplanarity_cost}
    \mathcal{E}_i(\bm{x}) \triangleq \frac{1}{2}\sum_j w_j \norm{\bm{\eta}^j_i(\bm{x})}^2_2, \ \ \bm{\eta}^j_i(\bm{x}) \triangleq \bm{n} \cdot\,\!^p\bm{p}_j - D
\end{gather}
where each $\bm{\eta}^j_i$ term represents the perpendicular distance of the $j$-th point $^p\bm{m}_j$ to the ground plane and $w_j \in \RR^+$ the weight. The ground plane is defined through the unit normal vector $\bm{n} \in \RR^3$ and the distance to the origin $D \in \RR^+$ (Hessian normal form). For convenience, we set the ground plane parameters to be $\bm{n} \triangleq (0, 0, 1)$ and $D \triangleq 0$ (see \secref{sec:coplanarity_relaxation_solution}).

The observed 3D incremental motions $(\bm{R}^k_i, \bm{t}^k_i) \in \text{SE(3)}$ for the $i$-th sensor can be projected to the ground plane using the estimated rotation matrix $\bm{R}_i$ from~\eqref{eq:coplanarity_ls} as:
\begin{equation}
    \bm{\bar{R}}^k_i \triangleq \bm{R}_i \bm{R}^k_i \bm{R}_i^\top,\quad \bm{\bar{t}}^k_i \triangleq \bm{R}_i \bm{t}^k_i,
\end{equation}
and then the SE(2) incremental motions can be easily recovered as the $x$-$y$ translation component of $\bm{\bar{t}}^k_i$, and by extracting the rotation angle about the $z$-axis from $\bm{\bar{R}}^k_i$.

These three parameters allow us to enforce the coplanarity constraint required by the motion-based calibration described in \secref{sec:two_coplanar_formulation} and, therefore, to estimate the full 6DoF extrinsic parameters.

%\begin{equation}
%    ^i\!\bm{\bar{R}}_k \approx R_z({p^k_i}_\theta)
%\end{equation}

%Finally, if the ground plane is observable, the remaining 3D transformation parameters, namely pitch and roll angles and the perpendicular distance, can be estimated. For 3D sensors like LIDAR or Depth sensors, these parameters are directly observable. For cameras, these additional parameters can be estimated from the homography decomposition, as described in~\cite{planar-odometry_calibration}.

%From \eqref{eq:relation_simplified}, 

%\begin{equation}
%    e_{ij}(\bm{x}) = \sum_k \norm{p^k_i - \ominus\, \mathbf{x} \oplus p^k_j \oplus \mathbf{x}}^2
%\end{equation}

%\begin{equation}
%    \mathbf{x} = \argmin_{\mathbf{x}} e_{ij}(\bm{x})
%\end{equation}

%let Sim(2) be the group of orientation-preserving similarity transformations in the 2D plane. The unknown extrinsic parameter we want to estimate $\bm{c} \in \text{Sim(2)}$.
%whose frame of reference coincides with the robot's frame of reference

%--------------------------------------------------------
%                   4. Method description
%--------------------------------------------------------

\section{Method Description} 
\label{sec:Method}

In the proposed method, we estimate the 6DoF calibration parameters in two main steps. First, for each sensor, we estimate the parameters of the relative rigid body transformation with respect to the ground plane. Then, we estimate the extrinsic 2D parameters for each sensor with respect to the reference one from coplanar incremental motions. More specifically, the calibration pipeline can be summarized as follows:
\begin{enumerate}
    \item Acquire sensor data while the robot is moving.
    \item Estimate per-sensor egomotion.
    \item Estimate \emph{z}, \emph{pitch} and \emph{roll} relative to the ground plane.
    \item Project estimated trajectories to the ground plane and resample synchronous incremental motions.
    \item Perform a motion-based calibration of the remaining parameters: \emph{x}, \emph{y}, and \emph{yaw}.
    \item Refine the initial 2D parameters  (\emph{x}, \emph{y}, \emph{yaw}) in a joint optimization framework.
\end{enumerate}

The rest of this section is structured as follows. In \secref{sec:two_coplanar_solution} we show how to solve the calibration of two coplanar sensors from incremental motions. Next, in \secref{sec:coplanarity_relaxation_solution}, we show how the parameters relative to the ground plane can be solved. Finally, in \secref{sec:practical_considerations} we discuss practical aspects, covering: the sampling of synchronous incremental motions, considerations for robust estimation, the final refinement, and the metric estimation of \emph{z} in the monocular case.

%--------------------------------------------------------
%             4.A Closed Form - Motion-based
%--------------------------------------------------------

\subsection{Closed Form Solution to the Motion-based Pair-wise Calibration Problem} \label{sec:two_coplanar_solution}

%The auto-initialization computes an initial estimate of the pursued parameters and, at the same time, discards outlier incremental motions. In this way, the initial guesses are computed by solving \eqref{eq:pairwise_ls} in closed form over the set of inliers. The next sections go into detail of this process.

%\subsubsection{Pair-wise Closed Form Solution}
In order to solve \eqref{eq:pairwise_ls}, we follow a similar approach as the one described in~\cite{odometry-sensor_calib}. The solution comes by first reducing the least-squares formulation to a quadratic system with a quadratic constraint. The constrained optimization problem is then uniquely solved in closed form by using the method of Lagrange multipliers.

To reduce the calibration problem to a quadratic system, we first rewrite the error terms in~\eqref{eq:error_terms_definition} as:
\begin{equation} \label{eq:pairwise_ls_error}
    \bm{\epsilon}^k_{ij} = \bm{x} \oplus \bm{p}^k_j - \bm{p}^k_i \oplus \bm{x}
\end{equation}

Next, we parameterize the calibration angle $x_\theta$ by two independent variables: $\cos x_\theta$ and $\sin x_\theta$. Grouping the unknown parameters into the vector $\bm{\varphi} \in \RR^5$
\begin{equation}
    \bm{\varphi} = \begin{bmatrix}{x_s}^{-1} & x_x & x_y & \cos x_\theta & \sin x_\theta \end{bmatrix}^\top
\end{equation}
we can write the error terms~\eqref{eq:pairwise_ls_error} as:
\begin{equation}
    \bm{\epsilon}_k = \bm{Q}_k \bm{\varphi},
\end{equation}
where the matrix $\bm{Q}_k$ contains the known coefficients:
\begin{equation}
    \bm{Q}_k \triangleq \begin{bmatrix}
    -{p^k_i}_x & 1 - \cos {p^k_i}_\theta & \sin {p^k_i}_\theta     & {p^k_j}_x & -{p^k_j}_y\\
    -{p^k_i}_y & -\sin {p^k_i}_\theta    & 1 - \cos {p^k_i}_\theta & {p^k_j}_y & {p^k_j}_x
    \end{bmatrix}.
\end{equation}

The cost function \eqref{eq:cost_definition} can be written compactly as:
\begin{equation}\label{eq:matrix_cost}
    \mathcal{C}_{ij}(\bm{x}) = \frac{1}{2}\bm{\varphi}^\top\bm{M}\bm{\varphi} + C,
\end{equation}
where $C \in \RR$ is a constant term and $\bm{M}$ the symmetric matrix:
\begin{equation} \label{eq:symmetric_m}
    \bm{M} \triangleq \sum_k {\bm{Q}_k}^\top\bm{Q}_k.
\end{equation}

Therefore, we can write the least squares problem \eqref{eq:pairwise_ls} equivalently as a quadratic system with a quadratic constraint:
\begin{equation} \label{eq:ls_quadratic}
%\begin{split}
    \bm{\varphi}^* = \argmin_{\bm{\varphi}} \bm{\varphi}^\top \bm{M} \bm{\varphi},\ \text{subject to}\ \varphi^2_4 + \varphi^2_5 = 1.
%\end{split}
\end{equation}

The constraint in \eqref{eq:ls_quadratic} corresponds to $\cos^2 x_\theta + \sin^2 x_\theta = 1$ and can be written in matrix form as:
\begin{equation}
    \bm{\varphi}^\top\bm{W}\bm{\varphi} = 1, \quad \text{with } \bm{W} \triangleq \begin{bmatrix} \bm{0}_{3\times3} & \bm{0}_{3\times2}\\
    \bm{0}_{2\times3} & \bm{I}_{2\times2}
    \end{bmatrix}.
\end{equation}

Thus, the constrained optimization problem \eqref{eq:ls_quadratic} can be solved considering the Lagrangian:
\begin{equation}
    \Lagr(\bm{\varphi}, \lambda) \triangleq \bm{\varphi}^\top\bm{M}\bm{\varphi} + \lambda\left(\bm{\varphi}^\top\bm{W}\bm{\varphi} - 1\right)
\end{equation}
and its necessary conditions for optimality:
\begin{equation}\label{eq:extrema_conditon}
    \frac{\partial\Lagr}{\partial\bm{\varphi}} = 2\bm{\varphi}^\top\left(\bm{M} + \lambda\bm{W}\right) = \bm{0}^\top.
\end{equation}

Since the scale factor $\varphi_1 \in \RR^+$ has to be a positive real number, the $5 \times 5$ matrix $\bm{M} + \lambda\bm{W}$ must be singular to satisfy \eqref{eq:extrema_conditon}. Thus, we solve for the value of $\lambda \in \RR$ that makes:
\begin{equation} \label{eq:lambda_condition}
    \det(\bm{M} + \lambda\bm{W}) = 0
\end{equation}
and then we find the solution $\bm{\varphi}^*$ in the kernel of $\bm{M} + \lambda\bm{W}$.

For the matrix $\bm{W}$, the expression \eqref{eq:lambda_condition} is a second order polynomial in $\lambda$ and, therefore, can be solved in closed form. The two candidate values $\lambda^*$ are computed as the real roots of the second-order polynomial. The rank of the $5 \times 5$ matrix $\bm{M} + \lambda^*\bm{W}$ is then at most 4 by construction. Provided that at least two linearly independent incremental motions are observed and that the two sensors follow nondegenerated trajectories, the matrix has rank exactly 4 and hence a one-dimensional kernel. For detailed information about the observability analysis we refer the reader to~\cite{coplanar_calibration, odometry-sensor_calib}.

To obtain the solution $\bm{\bar{\varphi}}^*$ associated to each $\lambda^*$, consider any nonzero vector $\bm{\gamma}^*$ in the kernel of $\bm{M} + \lambda^*\bm{W}$. Then, we impose the constraint \eqref{eq:ls_quadratic} and the fact that $\bar{\varphi}_1^*$ must be positive to uniquely identify $\bm{\bar{\varphi}}^*$ from the nullspace:
\begin{equation}
    \bm{\bar{\varphi}}^* = \frac{\sign(\gamma_1^*)}{\norm{\bigl[\begin{smallmatrix}\gamma_4^* & \gamma_5^*\end{smallmatrix}\bigr]^\top}_2}\bm{\gamma}^*.
\end{equation}

Then, we choose the correct solution $\bm{\varphi}^*$ to \eqref{eq:ls_quadratic} from the two $\bm{\bar{\varphi}}^*$ as the one that achieves the lowest cost according to~\eqref{eq:matrix_cost}. Finally, we recover the optimal calibration parameters $^i\bm{x}^*_j \in \text{Sim(2)}$ between the considered pair of sensors from $\bm{\varphi}^*$ as:
\begin{equation}
    ^i\bm{x}^*_j = \begin{bmatrix}\varphi^*_2 & \varphi^*_3 & \text{atan2}(\varphi^*_5, \varphi^*_4) & {\varphi^*_1}^{-1} \end{bmatrix}^\top.
\end{equation}

\subsection{Closed Form Solution to the Coplanarity Relaxation Problem}
\label{sec:coplanarity_relaxation_solution}

In order to solve~\eqref{eq:coplanarity_ls}, we again reduce the least-squares formulation to a quadratic system with a quadratic constraint. Then, the constrained optimization problem is uniquely solved in closed form by using the method of Lagrange multipliers.

To reduce the relaxation problem to a quadratic system, we first start by simplifying the error function~\eqref{eq:coplanarity_cost}. Recalling from the problem formulation, the Hessian normal parameters of the ground plane are:
\begin{equation}
    \bm{n} \triangleq \begin{bmatrix} 0 & 0 & 1\end{bmatrix}^\top,\quad D \triangleq 0
\end{equation}
and thus, the perpendicular distance $\bm{\eta}^j_i$ of the $j$-th point $^p\bm{m}_j$ to the ground plane reduces to its third component:
\begin{gather}
    \bm{\eta}^j_i = {^pm_j}_z = \bm{r}_z\,^i\bm{m}_j + x_z\\
    \bm{r}_z \triangleq \begin{bmatrix} -\sin x_\beta & \cos(x_\beta)\sin(x_\alpha) & \cos(x_\beta)\cos(x_\alpha) \end{bmatrix}
\end{gather}
where $\bm{r}_z$ represents the third row of the rotation matrix \mbox{$\bm{R} \in \text{SO(3)}$} as defined in~\eqref{eq:se3_parameters}, for the $i$-th sensor.

Therefore, we parameterize the $x_\alpha$ and $x_\beta$ angles by three independent variables: $-\sin x_\beta$, $\cos(x_\beta)\sin(x_\alpha)$ and $\cos(x_\beta)\cos(x_\alpha)$. Grouping the unknown parameters into the vector $\bm{\varphi} \in \RR^4$
\begin{equation}
    \bm{\varphi} = \begin{bmatrix} x_z & -\sin x_\beta & \cos(x_\beta)\sin(x_\alpha) & \cos(x_\beta)\cos(x_\alpha) \end{bmatrix}^\top
\end{equation}
we can write the error terms in matrix form
\begin{equation}
    \bm{\eta}^j = \bm{Q}_j\bm{\varphi},\quad \bm{Q}_j \triangleq \begin{bmatrix} 1 & {^im_j}_x & {^im_j}_y & {^im_j}_z \end{bmatrix},
\end{equation}
and, then, the cost function in~\eqref{eq:coplanarity_cost} yields:
\begin{equation}
    \mathcal{E}_i(\bm{x}) = \frac{1}{2}\bm{\varphi}^\top\bm{M}\bm{\varphi},
\end{equation}
with the symmetric matrix $\bm{M} \triangleq \sum_j w_j Q_j^\top Q_j$.

Therefore, we can write the least squares problem~\eqref{eq:coplanarity_ls} as the following quadratic system with a quadratic constraint:
\begin{equation} \label{eq:ls_quadratic2}
%\begin{split}
    \bm{\varphi}^* = \argmin_{\bm{\varphi}} \bm{\varphi}^\top \bm{M} \bm{\varphi},\ \text{subject to}\ \varphi^2_2 + \varphi^2_3 + \varphi^2_4 = 1.
%\end{split}
\end{equation}

The constraint in~\eqref{eq:ls_quadratic2} corresponds to the orthogonality property of SO(3) matrices and can be written as:
\begin{equation}
    \bm{\varphi}^\top\bm{W}\bm{\varphi} = 1, \quad \text{with } \bm{W} \triangleq \begin{bmatrix} 0 & \bm{0}_{1\times3}\\
    \bm{0}_{3\times1} & \bm{I}_{3\times3}
    \end{bmatrix}.
\end{equation}

For this particular matrix $\bm{W}$, the necessary conditions for optimality are characterized by a third order polynomial in $\lambda$ and therefore can be solved in closed form, with a maximum of three different candidate values $\lambda^*$. And as before, each solution $\bm{\bar{\varphi}}^*$ associated to a $\lambda^*$ can be uniquely recovered from any nonzero vector in the kernel of $\bm{M} + \lambda^*\bm{{W}}$ by imposing the orthogonality constraint~\eqref{eq:ls_quadratic2} and the fact that the perpendicular distance $\bar{\varphi}^*_1$ has to be positive.

The optimal, minimal solution $\bm{\varphi}^*$ to~\eqref{eq:ls_quadratic2} is chosen among all candidates $\bm{\bar{\varphi}}^*$ by evaluating the cost function~\eqref{eq:coplanarity_cost}. Finally, the calibration parameters for the $i$-th sensor $^i\bm{x} \in \RR^3$ easily follow from $\bm{\varphi}^*$.

%--------------------------------------------------------
%        4.C Considerations for Robust Calibration
%--------------------------------------------------------

\subsection{Practical Considerations}
\label{sec:practical_considerations}

First, the motion-based calibration procedure as described in~\secref{sec:two_coplanar_solution} requires synchronous incremental motions. However, heterogeneous sensors in general are asynchronous and have different sampling rates. We overcome this problem by setting a sensor to provide the time reference and then, for the other sensors, we resample synchronous incremental motions computed from linearly interpolated poses of the planar trajectories.

Secondly, motion-based calibration methods rely on trajectories estimated by other means. However, motion estimation algorithms are subject to large errors arising from local tracking failure or drift. In order to limit the impact that erroneous observations have on the calibration, a \emph{Random Sample Consensus} (RANSAC)~\cite{ransac_survey} framework can be used with our closed form solution. A RANSAC scheme requires a predefined threshold on the error terms in order to discard outliers. However, the error function as defined in~\eqref{eq:error_terms_definition} mixes the rotation error and the translation error, which have different magnitudes. What's more, the translation error is defined in the space of the second sensor (denoted by $j$ in~\eqref{eq:error_terms_definition}), and has an arbitrary scale in the case of a monocular camera. Therefore, we use a slightly different error function for the RANSAC outlier detection step:
\begin{equation} \label{eq:error_terms_fixed}
    \bm{\gamma}^k_{ij}(\bm{x}) \triangleq \trans(\bm{p}^k_i - \bm{x} \oplus \bm{p}^k_j \ominus\, \bm{x}),
\end{equation}
which is the translation error expressed in the metric space of the $i$-th sensor. The error terms defined in this way depend on the translation as well as on the rotation parameters and allow us to set a predefined threshold intuitively, with a simple interpretation. Note that the error function defined in~\eqref{eq:error_terms_definition} is still used for the closed form solution, and \eqref{eq:error_terms_fixed} is used only for the inlier-outlier classification in a RANSAC framework.

Another important consideration for practical calibration of multiple sensors is the consistency of the transformations between all sensors. The motion-based calibration presented in~\secref{sec:two_coplanar_solution} only considers constraints between a sensor and the reference one. We can improve the consistency of the calibration by also considering constraints between the other sensors in a joint optimization framework.

Let the reference sensor have index 0. Assuming we consider $n$ sensors for calibration, we want to estimate the calibration parameters $^0\bm{x}_1, \dots ^0\bm{x}_n \in \text{Sim(2)}$. We define the joint calibration problem as:
%\vspace{-0.2cm}
\begin{multline} \label{eq:joint_ls}
    ^0\bm{x}^*_1, \dots, ^0\bm{x}^*_n = \argmin_{^0\bm{x}_1, \dots, ^0\bm{x}_n}\, \sum^n_{i=1} \sum_k \norm{\rho\big(\bm{\gamma}^k_{0i}(^0\bm{x}_i)\big)}^2_2\\
    + \sum_{(i,j) \in S} \sum_k \norm{\rho\big(\bm{\gamma}^k_{ij}(\ominus\, ^0\bm{x}_i\oplus^0\!\bm{x}_j)\big)}^2_2,
\end{multline}
where $S$ represents the set containing all sensor pairs for which additional constraints are considered. The modified error function in~\eqref{eq:error_terms_fixed} is used again to express the error terms of all sensors in the same metric space. Note that only pairs of sensors that have their first sensor providing metrically accurate motions may be considered for additional constraints. The expression $\ominus\, ^0\bm{x}_i\oplus^0\!\bm{x}_j$ is just the relative transformation $^i\bm{x}_j$ expressed in terms of the calibration parameters $^0\bm{x}_i, ^0\bm{x}_j$. In our implementation, we used the Cauchy loss function $\rho$ to cope with unmodeled errors not detected during the RANSAC step. The joint calibration problem in~\eqref{eq:joint_ls} can be solved iteratively, starting from the closed form solution described in~\secref{sec:two_coplanar_solution}. For more information on how to solve this optimization problem, we refer the interested reader to~\cite{g2o}.

Finally, in the monocular camera case, the metric value of $x_z$ can be recovered by applying the scale factor estimated in the motion-based calibration to the $x_z$ value observed during the coplanarity relaxation. However, the same reconstruction has to be used for both problems in order to guarantee that they share the same scale.

\section{Experimental Evaluation} \label{sec:Experiments}
We have conducted several experiments in order to demonstrate the suitability of our calibration method in practice. First, we validated our method with simulated data, where the ground truth parameters are known (\secref{sec:validation_eval}). Next, we assessed the calibration accuracy of our method in both outdoor (\secref{sec:outdoor_eval}) and indoor (\secref{sec:indoor_eval}) environments with real data, and compared it with the state-of-the-art motion-based calibration approach proposed by Della Corte~\etal~\cite{motion-based_calibration_w_time}.

%--------------------------------------------------------
%              5.A 
%--------------------------------------------------------

\subsection{Validation with synthetic data} \label{sec:validation_eval}
The goal of the simulation experiment is to characterize the accuracy of our method. Here, the calibration parameters are known and the sensors are perfectly synchronized. We considered a two-sensor setting for the validation: an odometer and a monocular camera. While the odometer provides metrically accurate motions, the camera yields motions in a different scale (imitating the scale ambiguity problem).

In the simulation, we commanded the robot to follow an eight-shaped path several times. The incremental motions are affected by unbiased, normally distributed noise in each axis independently. We considered different levels of noise by varying the standard deviation: $\lambda \SI{0.1}{\centi\meter}$ for translation and $\lambda \SI{0.03}{\radian}$ for rotation, given a noise level $\lambda \in \RR^+$. The odometer is affected by noise only in the $x$-$y$ axes for translation and in the $z$ axis for rotation, since it provides 2D motion estimates. On the other hand, the camera motion has noise in the three axes (for translation and rotation).

The simulated camera has a QVGA resolution and a diagonal FOV of \SI{70.1}{\degree} with square pixels. For the coplanarity relaxation, the camera estimates a dense, scaled reconstruction of the ground plane in the same scale as the incremental motions. We added unbiased, normally distributed noise with $\lambda \SI{1}{\centi\meter}$ standard deviation to the depth measurements (for a noise level $\lambda$).

We considered 10 independent runs for different levels of noise. The relative transformation between the sensors is fixed and shared by all runs. The estimated parameters, as well as the ground truth are listed in \tabref{tab:simulation_results}. We also reported the average and the absolute and relative Root Mean Squared Error (RMSE). We included the noise-free case ($\lambda = 0$) to show the correctness of our formulation and compared it with the method in \cite{motion-based_calibration_w_time}. The approach in \cite{motion-based_calibration_w_time} incorrectly detects a time delay and bias its solution. Additionally, it cannot handle higher levels of noise. On the other hand, our method retrieves the correct solution in the noise-free case and is more robust under noisy observations. From the table we can also conclude that the calibration parameters estimated from the coplanarity relaxation are more accurate than the motion-based ones. This is mainly due to a higher number of observations (\num{76800} \emph{vs} \num{74}). Finally, the calibration errors are concentrated in the 2D translation parameters, since they depend on the estimation of all the other parameters (apart from $x_z$).

\tabsimulation

%--------------------------------------------------------
%              5.B
%--------------------------------------------------------

\subsection{Outdoor evaluation} \label{sec:outdoor_eval}

For the outdoor experimental evaluation, we chose the publicly available dataset in Blanco \etal~\cite{dataset_cm_accuracy_groundtruth}, as it provides highly accurate trajectories for different rigidly mounted sensors (including cameras) and ground truth extrinsic calibrations. The dataset was recorded with an electric buggy while driving in both planar and nonplanar surfaces. The vehicle's motion was estimated from a total of three Real Time Kinematics (RTK) GPS receivers and the ground truth extrinsic calibrations were initialized from manual measurements and then refined via nonlinear optimization techniques.

We selected the vehicle's trajectory as the reference motion and used the trajectory of the left camera as provided by the dataset for extrinsic calibration. Note that even though the sensor being calibrated is a camera, we are using the trajectory provided by the dataset, which is metrically accurate. The vehicle's 6DoF poses were recorded at \SI{1}{\hertz}, while the images were captured with $\num{1024}\times\SI{768}{\px}$ resolution at \SI{7.5}{\hertz}.

As our method requires synchronous incremental motions, we synchronized the trajectory of the camera with the vehicle's trajectory by linear interpolation. We chose such a trajectory as the synchronization time reference as it has the lower rate of the two and thus interpolated camera poses can be computed more accurately. Both the proposed method and the motion-based approach in~\cite{motion-based_calibration_w_time} are given the same three input trajectories for the sake of fairness. %In addition, we enabled the time delay estimation feature for the method in~\cite{motion-based_calibration_w_time}, since image timestamps are assigned by software when received by the computer and thus are subject to both acquisition and transmission time delays.

The calibration results of both methods for the three planar sequences are presented in \tabref{tab:outdoor_results}. For our method, the 3DoF calibration parameters of the relative transformation to the ground plane are estimated only once for all sequences (that is why no standard deviation values for these estimations are shown in the table). The 3D ground points are extracted from images by first running the Structure-from-Motion (SfM) pipeline~\cite{colmap} on 50 of them with enough texture on the ground and then detecting planar points on the reconstructed 3D scene through homographies. We used the metrically accurate trajectory of the camera to compute the scale factor in order to retrieve the metric value of $x_z$. That parameter is missing for [8], since it is unobservable solely from planar motions \cite{analytical_odometer-camera_calibration}. In general, both methods provide consistent parameters estimation. However, the method in~\cite{motion-based_calibration_w_time} required a close initial guess of the translation component in order to provide reasonable results.
%Our method underestimates the $x_x$ parameter, while the time delay 
%\red{However, small biases are affecting the \emph{x} component of the calibration parameters in our method. These deviations can be produced by a time delay on the camera, which does not affect the method~\cite{motion-based_calibration_w_time} as it is estimated along with the calibration parameters}.

\tabcalibrationsdataset

%--------------------------------------------------------
%              5.C
%--------------------------------------------------------

\subsection{Indoor evaluation} \label{sec:indoor_eval}
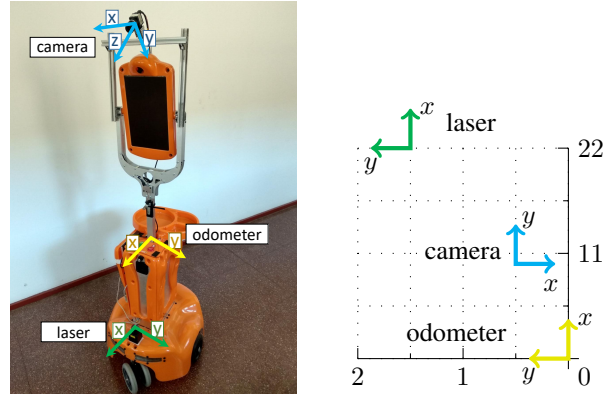
\begin{figure}[tb]
    \centering
    \includegraphics[width=0.21\textwidth]{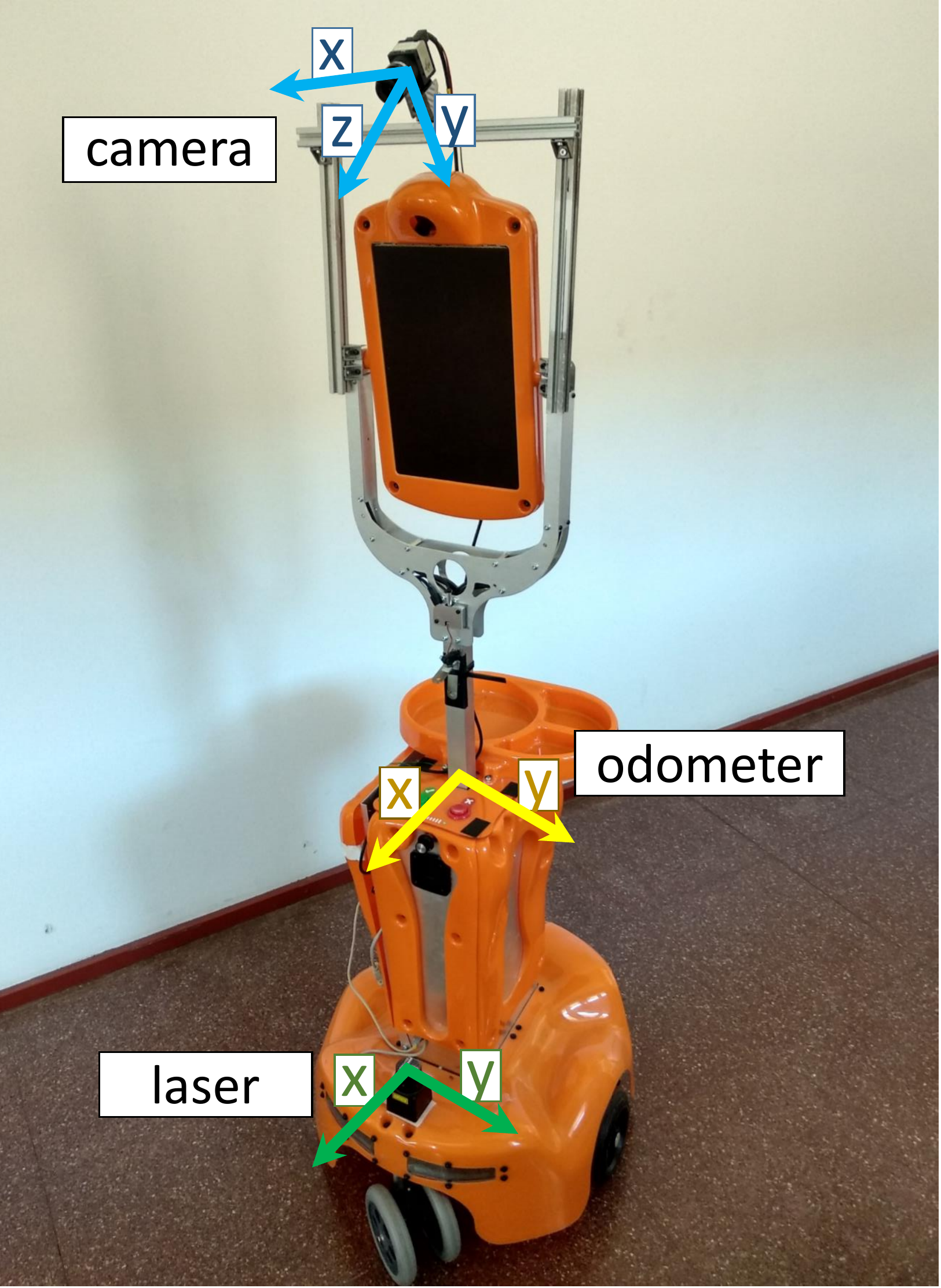}
    \hspace{1em}
    \begin{tikzpicture}[scale=0.7]
    \definecolor{odometer}{RGB}{230,230,0}
    \definecolor{laser}{RGB}{0,176,80}
    \definecolor{camera}{RGB}{0,176,240}
    %Reference systems
    \draw[loosely dotted] (0,0) grid (-4,4);
    % Axes
    \draw (0,0) -- (0,4);
    \draw (0,0) -- (0,-0.2);
    \draw (0,1) -- (-0.1,1);
    \draw (0,2) -- (-0.2,2);
    \draw (0,3) -- (-0.1,3);
    \draw (0,4) -- (-0.2,4);
    
    \draw (0,0) -- (-4,0);
    \draw (0,0) -- (0.2,0);
    \draw (-1,0) -- (-1,0.1);
    \draw (-2,0) -- (-2,0.2);
    \draw (-3,0) -- (-3,0.1);
    \draw (-4,0) -- (-4,0.2);
    % Odometer
    \draw [<->, ultra thick, odometer] (-0.75,0) -- (0,0) -- (0,0.75);
    \draw [<->, ultra thick, camera] (-0.25,1.8) -- (-1,1.8) -- (-1,2.55);
    \draw [<->, ultra thick, laser] (-3.75,4) -- (-3,4) -- (-3,4.75);
    %Labels
    \node [right] at (0,0.75) {$x$};
    \node [below] at (-0.75,0) {$y$};
    \node [left] at (-1, 0.5) {odometer};
    
    \node [right] at (-0.65,1.4) {$x$};
    \node [above] at (-0.75,2.3) {$y$};
    \node [left] at (-1.1, 2) {camera};
    
    \node [right] at (-3,4.75) {$x$};
    \node [below] at (-3.75,4) {$y$};
    \node [right] at (-2.5,4.5) {laser};
    %Ticks
    \node [below right] at (0,0) {\num{0}};
    
    \node [right] at (0,2) {\num{11}};
    \node [right] at (0,4) {\num{22}};
    
    \node [below] at (-2,0) {\num{1}};
    \node [below] at (-4,0) {\num{2}};
    \end{tikzpicture}
    \caption{Sensor setup for experimental evaluation. Left, Giraff robot equipped with an odometer, a laser scanner, and a camera. Right, schematic of the local reference systems projected on the ground plane (in \si{\centi\meter}).}
    \label{fig:giraff_config}
\end{figure}

For the indoor experimental evaluation, we used the Giraff mobile robot~\cite{raul2017ijrr,orlandini2016excite} equipped with an odometer (by integrating wheel encoders for a differential drive configuration), an Hokuyo UTW-30LX 2D laser scanner and an uEye UI-1240SE-M-GL monocular camera. The laser scanner is mounted in parallel to the ground, while the camera is pointing downwards with an incidence angle of about \SI{40}{\degree}, as depicted in \figref{fig:giraff_config}~(left). %The fact that the laser is mounted horizontally allows us to use the solution from \secref{sec:two_coplanar_solution} directly, since it is coplanar with our sensor of reference: the odometer. 
The laser scanner and the camera are rigidly connected to the robot and, therefore, the relative transformation between them does not change during the experiments. Both the odometer and the camera have been intrinsically calibrated before attempting extrinsic calibration.
% The sensors' measurements are asynchronous, but they are captured by the robot with the same time reference and, therefore, temporal offsets are assumed to be negligible.

%Since we do not have the ground truth of the relative transformations, 
We considered five independent calibrations and then we analyzed the mean and deviation of the estimated parameters. To this end, we recorded three data sequences with software acquisition timestamps, while driving the robot through our lab following an eight-shaped path several times. The incremental motions of the laser scanner are estimated using the method in~\cite{srf2o}, which provides 2D pose estimates at about \SI{6}{\hertz}. For the camera, we run ORB-SLAM~\cite{orb_slam} with $\num{1280}\times\SI{1024}{\px}$ images at \SI{5}{\hertz}.
%Both wheel and laser-scanner odometries are in 2D, while the outcome of the SfM are 3D poses, which are projected to the ground plane to recover the 2D motions.
%\red{The methods used for the odometer and the laser scanner directly provide 2D egomotions. For the camera, 3D poses are provided and we compute the 2D motions by projecting them into the floor plane, in which the other sensors' motions are expressed. The measurements of each sensor are captured by the robot with the same time reference.}

This time we synchronized the incremental motions of the laser with respect to the keyframes as selected by the SLAM solution. We chose the camera as the synchronization time reference in order to interpolate between odometry and laser poses, which are available at higher and more consistent frame rates. Each data sequence contains \numrange{70}{90} synchronized incremental motions.

%The ground plane cannot be observed with the radial laser scanner in the current configuration and, thus, the coplanarity relaxation presented in \secref{sec:coplanarity_relaxation_solution} is not applicable. However, the sensor's egomotion is already expressed in the ground plane since the laser is mounted horizontally. This is important because horizontal laser scanners are very common in mobile robotic platforms, being the motion-based calibration presented in \secref{sec:two_coplanar_solution} directly applicable in this configuration.

The experimental results within indoor environments for the considered methods are presented in \tabref{tab:indoor_results}, as well as the expected parameters for reference (from manual measurements). The robotic platform uses the laser for localization and thus contains rough calibration parameters for it. The motion estimation algorithm for the laser uses them to estimate the pose of the robot instead of the laser one. Thus, even though there is a translation of \SI{22}{\centi\meter} in the $x$ axis, the expected calibration parameter is $x_x = \SI{0}{\centi\meter}$. The $x_z$, $x_\beta$ and $x_\alpha$ parameters of the laser are omitted in \tabref{tab:indoor_results}, since they are unobservable with the current setup~\cite{per-sensor_egomotion_calibration}. Therefore, in the case of the laser scanner, we skipped the coplanarity relaxation for our method, and for \cite{motion-based_calibration_w_time} we locked the affected parameters to zero. As before, we estimated the 3DoF parameters of the camera relative to the ground plane only once for all sequences for our method.
The method in~\cite{motion-based_calibration_w_time} requires metrically accurate trajectories. Therefore, we applied the scale factor estimated by our method to the 3D trajectories estimated with ORB-SLAM. The results are similar to the ones from the simulation experiment (\secref{sec:validation_eval}). Both methods agree on the $x_y$ and $x_\theta$ parameters, while there are slight discrepancies in the other parameters. Looking at the deviations, both methods are consistent with their estimates. We argue that these differences are due to the time delay estimation in~\cite{motion-based_calibration_w_time}. Overall, our method provides calibration parameters closer to the measured ones.

\tabcalibrations

%--------------------------------------------------------
%                  6. Conclusions 
%--------------------------------------------------------

\section{Conclusions} \label{sec:Conclusion}

In this paper, we presented a new method for the extrinsic calibration of multiple sensors, suitable for automatic execution on mobile robots. In particular, we first formulated a least-squares problem to estimate the 2D calibration parameters of two coplanar sensors from incremental motions. Next, we relaxed the coplanarity requirement by first estimating the 3DoF transformation relative to the ground plane, and then projecting planar motions into the common plane. Then, we extended the two-sensor case by considering all sensor pairs in a joint least-squares framework. Our formulation allows to accurately estimate the 6DoF calibration parameters of multiple heterogeneous sensors under the assumptions that: i) the ground plane can be observed and ii) accurate per-sensor motions are available. A scale factor is also considered as an estimation parameter during the motion-based calibration and, therefore, our method can also handle monocular cameras. Finally, the proposed approach has been validated with simulated data and assessed in both indoor and outdoor scenarios, supporting its practical application and enhancing the performance of a state-of-the-art motion-based approach.
% Omitted for space constraints!
%The calibration framework has been made publicly available for the research community at~\url{https://github.com/dzunigan/robot_autocalibration}.
Currently, we are investigating closed form solutions to estimate sensors' time delays.

%\section*{Acknowledgment}
\vspace{2em}

% Bibliography
\bibliographystyle{IEEEtran}
\bibliography{IEEEabrv,ref}

\end{document}

%% file: tables.tex
\newcommand\tstrut{\rule{0pt}{2.6ex}}       % Top strut
\newcommand\bstrut{\rule[-1.5ex]{0pt}{0pt}} % Bottom strut

\newcommand\tabsimulation{
\begin{table}[tb]
\begin{center}
    \small\addtolength{\tabcolsep}{-4pt}
    \caption{Estimated calibration parameters w.r.t. the odometer}
    \label{tab:simulation_results}
    \scriptsize
    \begin{tabular}{|c|l||r r r r r r r||}
    \hline
    
    \multicolumn{2}{|c||}{Noise} & \multicolumn{1}{c}{$x_x$} & $x_y$ & \multicolumn{1}{c}{$x_z$} & \multicolumn{1}{c}{$x_\theta$} & \multicolumn{1}{c}{$x_\beta$} & \multicolumn{1}{c}{$x_\alpha$} & \multicolumn{1}{c||}{\tstrut $x_s$ \bstrut} \\ \hline \hline
    
    \multicolumn{2}{|c||}{Ground truth} & \tstrut \SI{50}{\centi\meter} & \SI{10}{\centi\meter} & \SI{100}{\centi\meter} & \SI{-90}{\degree} & \SI{4.77}{\degree} & \SI{-135}{\degree} & \bstrut \num{2} \\ \hline \hline
    
   \multirow{2}{*}{$\lambda = 0$} & Avg. & \textbf{\SI[detect-weight]{50}{\centi\meter}} & \textbf{\SI[detect-weight]{10}{\centi\meter}} & \textbf{\SI[detect-weight]{100}{\centi\meter}}  & \textbf{\SI[detect-weight]{-90}{\degree}} & \textbf{\SI[detect-weight]{4.77}{\degree}} & \textbf{\SI[detect-weight]{-135}{\degree}} & \tstrut \bstrut \textbf{2} \\ \cline{2-9}
   & \multicolumn{1}{c||}{\cite{motion-based_calibration_w_time}} & \SI{44.6}{\centi\meter} & \textbf{\SI[detect-weight]{10}{\centi\meter}} & -- & \SI{-95.2}{\degree} & \SI{-12.2}{\degree}  & \SI{-150.9}{\degree} & \tstrut \bstrut -- \\ \hline \hline
    & Avg. & \SI{49.2}{\centi\meter} & \SI{9.8}{\centi\meter} & \SI{99.5}{\centi\meter}  & \SI{-89.8}{\degree} & \SI{4.77}{\degree} & \SI{-135}{\degree} & \tstrut \bstrut \num{1.99} \\ \cline{2-9}
     $\lambda = 1$ & Abs. & \SI{1.0}{\centi\meter} & \SI{0.2}{\centi\meter} & \SI{0.5}{\centi\meter} & \SI{0.5}{\degree} & \SI{0.0}{\degree}  & \SI{0.01}{\degree} & \tstrut \bstrut \num{0.01} \\ \cline{2-9}
    & Rel. & \SI{2.15}{\percent} & \SI{2.45}{\percent} & \SI{0.6}{\percent} & \SI{0.6}{\percent} & \SI{0.0}{\percent}  & \SI{0.01}{\percent} & \tstrut \bstrut \SI{0.6}{\percent} \\ \hline \hline
    
    & Avg. & \SI{46.8}{\centi\meter} & \SI{9.4}{\centi\meter} & \SI{98.4}{\centi\meter}  & \SI{-90.27}{\degree} & \SI{4.77}{\degree} & \SI{-134.9}{\degree} & \tstrut \bstrut \num{1.97} \\ \cline{2-9}
     $\lambda = 2$ & Abs. & \SI{3.4}{\centi\meter} & \SI{0.7}{\centi\meter} & \SI{1.6}{\centi\meter} & \SI{0.7}{\degree} & \SI{0.0}{\degree}  & \SI{0.04}{\degree} & \tstrut \bstrut \num{0.03} \\ \cline{2-9}
    & Rel. & \SI{6.9}{\percent} & \SI{7.1}{\percent} & \SI{1.6}{\percent} & \SI{0.8}{\percent} & \SI{0.0}{\percent}  & \SI{0.03}{\percent} & \tstrut \bstrut \SI{1.5}{\percent} \\ \hline
    
    \end{tabular}
\end{center}
\end{table}
}

\newcommand\tabcalibrationsdataset{
\begin{table}[tb]
\begin{center}
    \small\addtolength{\tabcolsep}{-4pt}
    \caption{Estimated calibration parameters w.r.t. the vehicle (translations are in~\si{\meter} and rotations in degrees)}
    \label{tab:outdoor_results}
    \scriptsize
    \begin{tabular}{|c|l||r r r r r r||}
    \hline
    
    \multicolumn{2}{|c||}{\multirow{2}{*}{\tstrut Calibration \bstrut}} & \multicolumn{6}{c||}{\tstrut Camera} \\
        
    \cline{3-8} \multicolumn{2}{|c||}{} & \tstrut $x_x$ & $x_y$ & $x_z$ & $x_\theta$ & $x_\beta$ & \bstrut $x_\alpha$ \\ \hline \hline
    
    \multicolumn{2}{|c||}{Ground truth} & \num{2.21} & \num{0.43} & \num{2.25} & \num{-88.4} & \num{-3.0} & \tstrut \bstrut \num{-87.2} \\ \hline \hline
    
    & \texttt{parkin\_0L} & \num{2.19} & \num{0.43} & \num{2.37}  & \num{-88.4} & \num{-3.7}  & \tstrut \num{-87.8} \\
    & \texttt{parkin\_2L} & \num{2.20} & \num{0.43} & \num{2.37}  & \num{-88.4} & \num{-3.7}  & \tstrut \num{-87.8} \\
    Ours & \texttt{parkin\_6L} & \num{2.19} & \num{0.43} & \num{2.37}  & \num{-88.4} & \num{-3.7}  & \tstrut \bstrut \num{-87.8} \\ 
    \cline{2-8}
    & Avg. & \num{2.20} & \textbf{\num[detect-weight]{0.43}} & \textbf{\num[detect-weight]{2.37}} & \textbf{\num[detect-weight]{-88.4}} & \num{-3.7} & \tstrut \bstrut \num{-87.8} \\
    \cline{2-8}
    \multicolumn{1}{|c|}{} & Stdv. & $\bm{0.01}$ & $\bm{0.00}$ & -- & $\bm{0.1}$ & -- & \tstrut \bstrut -- \\ \hline \hline
    
    & \texttt{parkin\_0L} & \num{2.20} & \num{0.42} & -- & \num{-87.5} & \num{-2.8}  & \tstrut \num{-87.3} \\
    & \texttt{parkin\_2L} & \num{2.21} & \num{0.39} & -- & \num{-90.3} & \num{-3.0}  & \tstrut \num{-87.1} \\
    Della Corte & \texttt{parkin\_6L} & \num{2.22} & \num{0.37} & -- & \num{-93.2} & \num{-0.7}  & \tstrut \bstrut \num{-87.0} \\ \cline{2-8}
    \etal~\cite{motion-based_calibration_w_time} & Avg. & 
    \textbf{\num[detect-weight]{2.21}} & \num{0.39} & -- & \num{-89.5} & \textbf{\num[detect-weight]{-3.0}} & \tstrut \bstrut \textbf{\num[detect-weight]{-87.1}} \\ \cline{2-8}
    \multicolumn{1}{|c|}{} & Stdv. & $\bm{0.01}$ & \num{0.03} & -- & \num{1.76} & \num{0.21} & \tstrut \bstrut \num{0.15} \\ \hline

    \end{tabular}
\end{center}
\end{table}
}

\newcommand\tabcalibrations{
\begin{table}[tb]
\begin{center}
    \small\addtolength{\tabcolsep}{-4.5pt}
    \caption{Estimated calibration parameters w.r.t. the odometer (translations are in~\si{\centi\meter} and rotations in degrees)}
    \label{tab:indoor_results}
    \scriptsize
    \begin{tabular}{|c|l||r r r |r r r r r r||}
    \hline
    \multicolumn{2}{|c||}{\multirow{2}{*}{\tstrut Calibration \bstrut}} & \multicolumn{3}{c|}{\tstrut Laser scanner} & \multicolumn{6}{c||}{\tstrut Monocular Camera} \\

    \cline{3-11} \multicolumn{2}{|c||}{} & \tstrut $x_x$ & $x_y$ & $x_\theta$ & \bstrut  \tstrut $x_x$ & $x_y$ & $x_z$ & $x_\theta$ & $x_\beta$ & \bstrut $x_\alpha$ \\ \hline \hline
    
    \multicolumn{2}{|c||}{Expected} & \num{0.0} & \num{0.0} & \num{0.0} & \num{10.0} & \num{0.5} & \num{175} & \num{-95} & \num{2.5} & \tstrut \bstrut \num{-130} \\ \hline \hline
    
    & \#1  & \num{-0.4} & \num{1.2} & \num{-3.0}  & \num{11.1} & \num{0.5}  & \num{175.6} & \num{-93.7} & \num{3.85} & \tstrut \num{-125.4} \\
    & \#2  & \num{-0.3} & \num{1.7} & \num{-2.3} & \num{11.0} & \num{0.2}  & \num{175.6} & \num{-93.6} & \num{3.85} & \num{-125.4} \\
    Ours & \#3  & \num{0.0} & \num{1.1} & \num{-2.9} & \num{11.6} & \num{0.4}  & \num{175.6} & \num{-93.9} & \num{3.85} & \num{-125.4} \\ 
    
    \cline{2-11}
    & Avg. & \textbf{\num[detect-weight]{-0.2}} & \textbf{\num[detect-weight]{1.3}} & \textbf{\num[detect-weight]{-2.8}} & \textbf{\num[detect-weight]{11.3}} & \textbf{\num[detect-weight]{0.4}} & \textbf{\num[detect-weight]{175.6}} & \textbf{\num[detect-weight]{-93.8}} & $\bm{3.85}$ & \tstrut \bstrut $\bm{-125.4}$  
    
    \\ \cline{2-11}
    \multicolumn{1}{|c|}{} & Stdv. & \textbf{\num[detect-weight]{0.2}} & \textbf{\num[detect-weight]{0.3}} & \textbf{\num[detect-weight]{0.4}} & 
    \num{0.34} & 
    \textbf{\num[detect-weight]{0.15}} &
    -- &
    \textbf{\num[detect-weight]{0.13}} &
    -- &
    \tstrut \bstrut -- \\ \hline \hline
    
    \multicolumn{1}{|c|}{\multirow{2}{*}{\cite{motion-based_calibration_w_time}}} & Avg. & \num{-2.8} & \num{1.4} & \num{-2.9} & \num{8.4} & \textbf{\num[detect-weight]{0.4}} & -- & \num{-101.6} & \num{-17.2} & \tstrut \bstrut \num{-137.3} \\ 
    
    \cline{2-11}
    \multicolumn{1}{|c|}{} & Stdv. &
    \textbf{\num[detect-weight]{0.2}} & \num{0.5} & \num{0.6} & \textbf{\num[detect-weight]{0.33}} & \num{0.24} & -- & \num{0.5} & \num{0.07} & \tstrut \bstrut \num{0.08} 
    
    \\ \hline
    \end{tabular}
\end{center}
\end{table}
}